\documentclass{svproc}
\usepackage{url}
\usepackage{mathtools}
\DeclarePairedDelimiter{\ceil}{\lceil}{\rceil}
\usepackage{amsmath}

\usepackage{graphicx}
\usepackage{times}
\usepackage{booktabs}
\usepackage{graphicx}
\usepackage{lipsum}
\usepackage{hyperref}       
\usepackage{url}      
\usepackage{breakurl}                                                          
\usepackage{booktabs}    
\usepackage{amsfonts}                                                    
\usepackage{nicefrac}                                                  
\usepackage{listings}       
\usepackage{amssymb}       
\usepackage{amsmath}   
\usepackage{eqnarray}
\usepackage{mathtools}  
\usepackage{lipsum}        
\usepackage{adjustbox}         
\usepackage{booktabs}     
\usepackage{multirow}

\usepackage{ifthen}       
\usepackage{color}      
\usepackage{xcolor}   
\usepackage[]{algorithm} 
\usepackage{subcaption}

\begin{document}
\mainmatter              
\title{PZnet: Efficient 3D ConvNet Inference on Manycore CPUs}
\titlerunning{PZnet: Efficient 3D ConvNet Inference on Manycore CPUs}  
%

\author{Sergiy Popovych\inst{1} \and Davit Buniatyan\inst{1} \and 
Aleksandar Zlateski\inst{2} \and Kai Li\inst{1} \and H. Sebastian Seung\inst{1}}
%
%
%
\institute{Princeton University, Princeton NJ 08544, USA,
\and
Massachusetts Institute of Technology, 77 Massachusetts Avenue, Cambridge, MA 02139, USA}

\maketitle              

\begin{abstract}
Convolutional nets have been shown to achieve state-of-the-art accuracy in many biomedical image analysis tasks. Many tasks within biomedical analysis domain involve analyzing volumetric (3D) data acquired by CT, MRI and Microscopy acquisition methods. To deploy convolutional nets in practical working systems, it is important to solve the efficient inference problem. Namely, one should be able to apply an already-trained convolutional network to many large images using limited computational resources. In this paper we present PZnet, a CPU-only engine that can be used to perform inference for a variety of 3D convolutional net architectures. PZNet outperforms MKL-based CPU implementations of PyTorch and Tensorflow by more than 3.5x for the popular U-net architecture. Moreover, for 3D convolutions with low featuremap numbers, cloud CPU inference with PZnet outperfroms cloud GPU inference in terms of cost efficiency.

\end{abstract}
\section{Introduction}
Convolutional neural networks (ConvNets) are becoming the primary choice of automated biomedical image analysis \cite{ronneberger2015u,seyedhosseini2013image,hariharan2015hypercolumns,lee2017dlinmed}, achieving superhuman accuracy for tasks such as chest X-ray anomaly detection\cite{rajpurkar2017chexnet}, neuron segmentation \cite{lee2017superhuman} and more\cite{madani2018echodl}. Many tasks within biomedical analysis domain involve analyzing volumetric data acquired by CT, MRI and Microscopy acquisition methods. As a result, tasks such as organ/substructure segmentation, object/lesion detection and exam classification  commonly employ 3D ConvNets \cite{litjens2017survey}.

Computational costs of ConvNet inference, which involves applying an already trained ConvNet to new images, is of a particular concern to the biomedical image analysis community. Each super resolution volumetric specimen can grow to the order of $10,000^3$ voxels \cite{tomer2014advanced,zheng2017complete}, resulting in vast amounts of data to be analyzed. Additionally, convolution operations often require more computation per pixel in 3D than 2D, which increases computational demand of 3D ConvNet inference. High computational costs of processing datasets can serve as a limiting factor in the quality of analysis. Thus, increasing utilization of available hardware resources for 3D ConvNet inference is a critical task. 

Modern deep learning frameworks such as Theano \cite{bastien2012theano}, Caffe \cite{jia2014caffe}, MXNet\cite{chen2015mxnet}, Tensorflow \cite{abadi2016tensorflow} and Pytorch \cite{collobert2011torch7} are mostly optimized for processing of 2D images, and achieve lower hardware utilization on both CPU and GPU platforms for 3D tasks. In this work we show that CPU efficiency for 3D ConvNet inference can be improved by up to 4x, which results in higher utility of existing CPU infrastructure and makes CPU inference a competitive choice in the cloud setting.
 
The main contribution of this work is an inference-only deep learning engine called PZnet, which is specifically optimized for 3D inference on Intel Xeon CPUs. PZnet utilizes ZnnPhi\cite{zlateski2017znnphi}, a state-of-the-art direct 3D convolution implementation. ZnnPhi relies on template-based metaprogramming and requires a custom data layout, which makes it not compatible with mainstream deep learning frameworks. For this reason, we created a special inference-only framework PZnet. A convolutional net can be trained using a mainstream deep learning framework, and then imported to PZnet when large-scale inference is required.  

PZnet implements a number of unique optimizations that complete operations required by several layers in one memory traversal. Reducing the number of memory traversals can be critical for achieving high performance on CPU platforms. These layer fusions are applicable to a number of ConvNet architectures, and can reduce inference time by up to 12\%. 

PZnet outperforms MKL-based \cite{intelcaffe} CPU implementations of PyTorch and Tensorflow by 3-8x, depending on the network architecture and hardware platform.  Moreover, we show that based on current cloud compute prices, PZnet CPU inference is competitive with cuDNN based GPU inference. For inference of a real-world residual 3D Unet architecture \cite{lee2017superhuman}, PZnet is able to outperform GPU inference in terms of cost efficiency by over $50\%$. To the best of our knowledge, this is the first work to show CPU inference to beat GPU inference in terms of cloud cost.

\section{PZnet}

\subsection{Problem Statement}
The goal of this work is to reduce the computational costs of running large scale 3D dense prediction tasks. We achieve this goal by maximizing efficiency of 3D ConvNet inference on CPUs. More specifically, we focus on Intel Xeon processors. A high efficiency CPU inference engine would improve the utility of existing CPU cluster infrastructure and reduce the costs spent on cloud resources. 

\subsection{Overview}
PZnet is a deep learning engine for 3D inference on Intel Xeon processors. PZnet achieves high efficiency through employing a specialized convolution implementation and performing a series of inter-layer optimizations. PZnet is compatible with the standard Caffe prototxt network specification format, which simplifies importing models trained in other frameworks. PZnet provides support for the following layers:

\begin{itemize}
  \item Convolution, strided and non-strided
  \item Deconvolution, strided and non-strided
  \item Batch Normalization
  \item Scale
  \item ReLU
  \item ELU
  \item Sigmoid
  \item Elementwise (Addition, Difision, Multiplication)
  \item MergeCrop
  \item Pooling (Average, Max)
\end{itemize}

PZnet consists of 2 parts -- a network generator and an inference API. Network generator compiles the provided model specifications into so-called \textit{network files}. Network files are shared library objects that are distributed to worker machines. Workers run inference by accessing the models within the network files through PZnet python inference API.

PZNet employs ZnnPhi, which, to the best of our knowledge, is the most efficient 3D direct convolution implementation known up to date. ZnnPhi achieves high performance though utilizing SIMD instructions in a cache efficient way, and is compatible with SSE4, AVX, AVX2 and AVX512 SIMD instruction families.

ZnnPhi requires image and kernel data to conform to a specific data layout. An image tensor with $B$ batches, $F$ featuremaps, and $X, Y, Z$ spacial dimensions has to be stored as an array with dimensions  $B \times \ceil{F/S} \times X \times Y \times Z \times S$, where S is the width of the SIMD unit. A kernel tensor of size $F' \times F \times K_x \times K_y \times K_z$ will be stored as an array with dimensions $\ceil{F'/S} \times \ceil{F/S} \times K_X \times K_Y \times K_Z \times S \times S$. ZnnPhi requires such data layout in order to maximize the efficiency of SIMD instruction utilization, and it prevents ZnnPhi from being pluggable into mainstream deep learning frameworks. 

Additionally, ZnnPhi heavily relies on metaprogramming through C++ templates, which means that layer parameters, such as image and kernel sizes, have to be known during compile time. This allows ZnnPhi to rely on compile time optimizations in order to produce maximally efficient code for each parameter configuration. However, this adds another obstacle to integrating ZnnPhi into a mainstream deep learning frameworks. Most deep learning frameworks allow user-supplied C++ layer implementations, but they either require them as a compiled shared object or as generic source code. Neither option is compatible with ZnnPhi, because ZnnPhi kernels need to be recompiled for each layer configuration.

In order to support ZnnPhi, PZnet compiler generates C++ source code which directly corresponds to the provided model. Then, Intel C++ Compiler is invoked in order to produce optimized shared library object files. All PZnet layers support ZnnPhi blocked memory layout. PZnet implicitly performs memory layout transformations for the input and output data tensors in order to provide standard input and output formats.

\section{Optimizations}
\begin{figure*}[h]
  \centering
  \includegraphics[width=340pt]{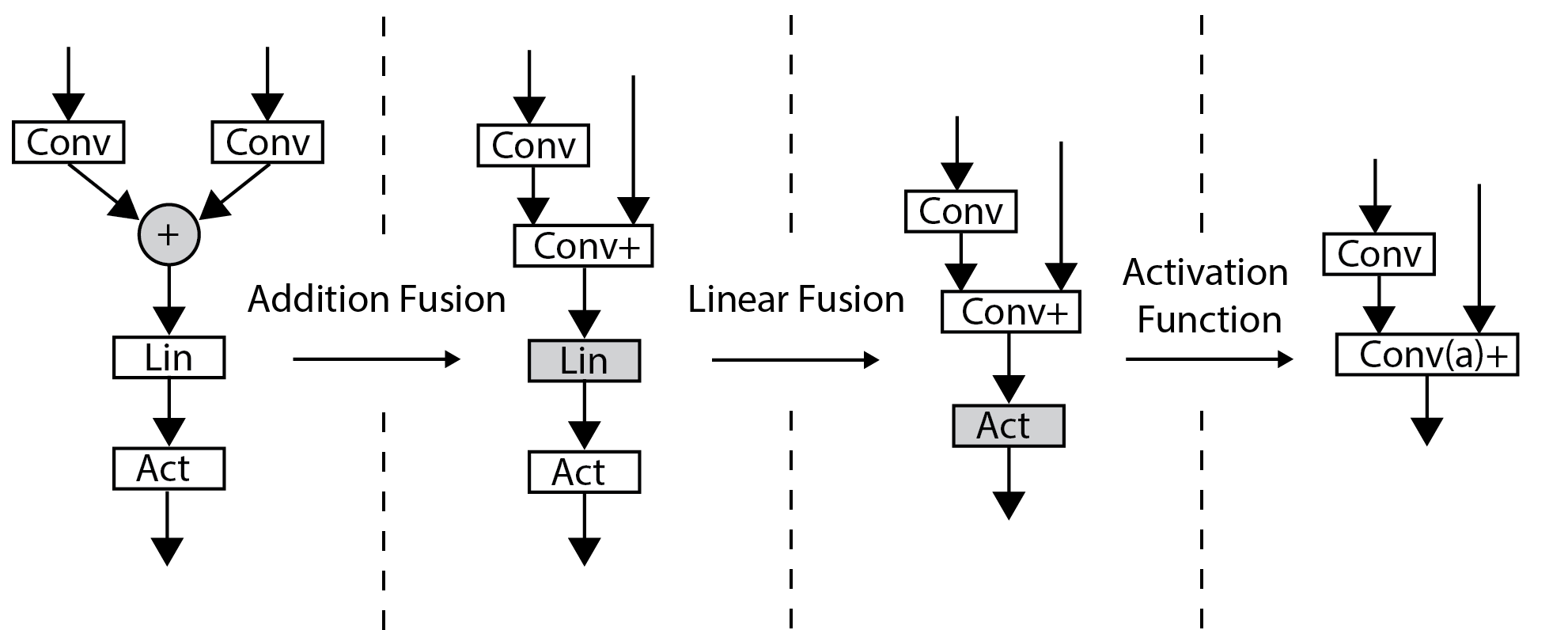}
  \caption{PZnet optimization flow}
  \label{opts}
\end{figure*}

Optimizations performed by PZnet are mainly aimed at reducing the number of memory traversals introduced by non-convolutional layers layers (batchnorm, scale, activation, etc). We modify ZnnPhi convolutional kernel in order to be able to perform several layer operations in one memory pass. The overall optimization flow of PZnet is shown on Fig \ref{opts}. 

First, we fuse convolution layers with element-wise addition layers, which are commonly introduced by residual connections. After element-wise layers are fused into convolutions, more convolution layers immediately precede batch normalization and scale layers (Fig \ref{opts}). During inference time, both batch normalization and scale layers perform linear transformations of tensors. Weights of the convolution layers can be modified in order to take account for subsequent linear transformation layers, and so we are able to fuse batch normalization and scale layers into the preceding convolution. After linear transformation layers are fused, convolutions are commonly followed by activation layers. We modify ZnnPhi primitives in order to apply activation function to the convolution outputs before they are written out to memory. Finally, after element-wise addition, linear transformation and activation layers are removed, most of convolution layer outputs are used inputs of the subsequent convolution layers. Thus, we can eliminate explicit input padding of the inputs by making convolution layers produce padded outputs, which saves additional memory traversals. Overall,  optimization by $7-12\%$ is performed, depending on CPU parameters and network architecture.

\subsection{Element-wise Addition Fusion}
State-of-the-art 3D segmentation models often include residual connections\cite{lee2017superhuman}, resulting in up to 1 element-wise addition layer per 3 convolution layers. We observe that we can eliminate these layers by modifying ZnnPhi convolution implementation. 

\begin{figure}[h]
  \centering
  
  \includegraphics[width=220pt]{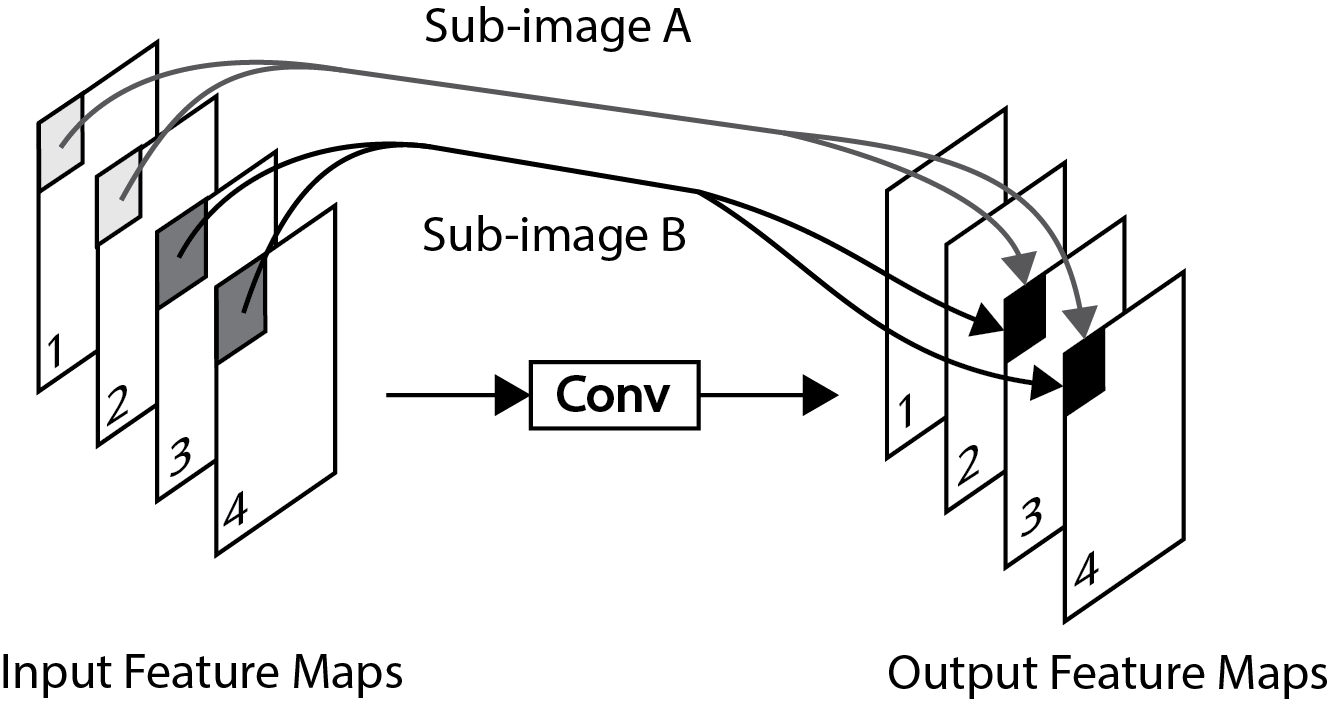}
  \caption{Application of two ZnnPhi sub-image primitives to the black patches in the output feature maps ($SIMD\_width = 2$). Each application computes the contribution of $SIMD\_width$ consecutive feature maps to the black output patch regions. After A and B are applied, the output patches will hold their final values. }
  \label{subimage}
\end{figure}

ZnnPhi achieves cache efficiency by implementing convolution as series of hierarchical primitives, with each higher level primitive utilizing the lower level ones in order to complete more complex tasks. The lowest level ZnnPhi primitive, called \textit{sub-image primitive}, will be extended in this work. The goal of the sub-image primitive is to compute the contribution S consecutive input feature maps to a small patch in the consecutive S output feature maps, where S is the SIMD width for the given instruction family. For example, Fig \ref{subimage} illustrates an application of 2 sub-image primitives, black and gray, for the case when $S = 2$. Both applications of the primitive target the same patches in output feature maps 3 and 4. The gray application initializes output patches to the bias values for the given output feature maps and adds to them the result of convolving kernels with patches in input feature maps 1 and 2. The black application finishes the computation by adding on the result of convolving kernels with patches in input feature maps 3 and 4. 

Sub-image primitive is designed to maximize register reuse and L1 cache hitrate. It is used by higher level primitives repeatedly in order to generate full output feature maps.

Element-wise Addition Fusion is performed as follows. Assume that element-wise addition layer E receives two input tensors which are produced by layers L1 and L2. Without loss of generality, we assume that L1 is a convolution layer. The following conditions need to be satisfied for fusion of E and L1. First, element-wise addition E has to be the only consumer of the output produced by convolution L1. Second, there must exist a topological ordering of the network graph in which L1 appears \textit{after} L2. If both of these conditions are satisfied, element-wise addition E can be fused into convolution L1. After the fusion, the tensor produced by L2 is passed to L1 as an additional \textit{Base} argument. Lastly, the additive flag for L1 is set to true. When the additive flag is set to true, the first application of the sub-image primitive initializes the output values to be the sum of the layer bias and the Base argument.

\subsection{Element-wise Addition Fusion}
Element-wise Addition Fusion is performed as follows. Assume that element-wise addition layer E receives two input tensors which are produced by layers L1 and L2. Without loss of generality, we assume that L1 is a convolution layer. The following conditions need to be satisfied for fusion of E and L1. First, element-wise addition E has to be the only consumer of the output produced by convolution L1. Second, there must exist a topological ordering of the network graph in which L1 appears \textit{after} L2. If both of these conditions are satisfied, element-wise addition E can be fused into convolution L1. After the fusion, the tensor produced by L2 is passed to L1 as an additional \textit{Base} argument. Lastly, the additive flag for L1 is set to true. When the additive flag is set to true, the first application of the sub-image primitive initializes the output values to be the sum of the layer bias and the Base argument.

\subsection{Linear Transformation Fusion}
Let's define a notation in which convolution layer takes an input of $N$ input feature maps and produces an output of $M$ feature maps, with each output feature map $O_{m}$ described as 
\begin{equation*}\label{eq:conv}
    O_{m} = \sum_{n=1}^{N} I_{n} \ast K_{nm} + B_{m} 
\end{equation*}
where $I_{n}$ denotes input feature map $n$, $K$ denotes convolution kernel and $B$ denotes bias. For layers that perform linear transformation during inference, such as batch normalization and scaling, the computation of each output feature map can be described as
\begin{equation*}\label{eq:linear}
    O_{m} = I_{m}\cdot M_{m} + A_{m} 
\end{equation*}
where $M_{m}$ and $A_{m}$ denote multiplicative and additive weights for feature map $m$. 

The combined computation of a convolution layer followed by a linear transformation layer can be described as
\begin{equation*}
\begin{aligned}\label{eq:convlin}
    O_{m} = (\sum_{n=1}^{N} I_{n} \ast K_{nm} + B_{m} ) \cdot M_{m} + A_{m} =\\= \sum_{n=1}^{N} I_{n} \ast (K_{nm} \cdot M_{m}) + (B_{m} \cdot M_{m} + A_{m})
\end{aligned}
\end{equation*}
which is equivalent to performing a convolution with kernel $K'_{nm} = K_{nm} \cdot M_{m}$ and bias $B'_{m} = B_{m} \cdot M_{m} + A_{m}$. Thus, we can eliminate linear transformation layers that immediately follow convolutions by performing the corresponding weights modifications. Note that this optimization applies only to inference, as presence of batch normalization and scale significantly affects network behavior during training.

\subsection{Activation Fusion}
In ZnnPhi the sub-image primitive is applied repeatedly to each output patch, with the final application storing the output patch values to memory. We observe that when convolution is immediately followed by activation, activation functions can be applied to output during the final application of sub-image primitive to each patch. This way the computed output values are already in cache during application activation, which eliminates memory traversal overhead caused by activation layers.

\subsection{Padding Transformation}

ZnnPhi does not support input padding, and so explicit padding layers have to be added to the IR of the network after the parsing phase. Padding cannot be done in-place, which means that memory traversals introduced by padding layers especially hurt inference efficiency. Moreover, due to specificity of the data layout required by ZnnPhi, the memory must be moved in a disjoint small junks in order to perform padding, which further hurts efficiency.

However, ZnnPhi allows the user to specify strides for each of the output dimensions. We observe that by manipulating the stride values for the spacial dimensions we can make ZnnPhi convolutions produce padded outputs. Thus, when two convolutions follow each other, the first convolution can produce output which is already padded for the second convolution. More formally, whenever all consumers of a convolution layer output require the same spacial padding, explicit input padding for the consumers can be avoided by generating padded output at the producer layer.

\begin{figure*}[h]
  \centering
  
  \includegraphics[width=340pt]{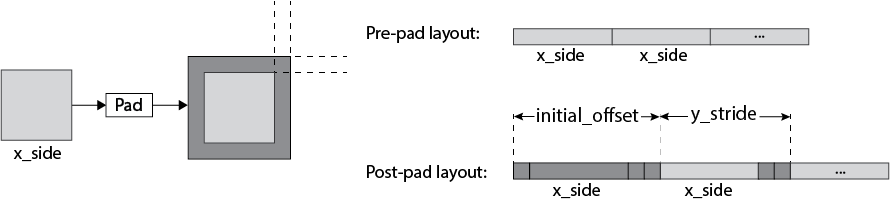}
  \caption{Left: operation of the padding layer. Right: memory layout of the image before and after padding. The padded output can be viewed as a representation of the input image with $initial\_offset = x\_side + 3$ and $y\_stride = x\_side + 2$}
  \label{padding}
\end{figure*}

Fig~\ref{padding} illustrates how manipulating initial offset and strides can be used in order to generate padded outputs for row-major memory layout. In the illustrated case, a 2D image is to be padded by $1$ in box $x$ and $y$ dimensions. This can be achieved by setting the initial offset to be $size(y) + 3$, and the stride for the y dimension to be $size(y) + 2$.  Analogously, padding can be generated for 3D images with ZnnPhi blocked memory layout. 

\section{Evaluation}

\subsection{Setup}

The experiments in this section are performed on major types of CPU and GPU compute instances from Amazon Web Services (AWS), namely c4, c5, p2 and p3 instance types. 

Processing of volumetric datasets is generally done by breaking the volume into a large number of smaller patches. The segmentation result of each patch is computed as a separate task. When all of the patches are computed, the results are concatinated to obtain the final result. Small task granularity minimizes distortions caused by instance termination, which makes spot instances a perfect choice of for large scale inference.

Small task granularity also encourages the use of small instances for execution of each individual task. It is easier to achieve high utilization on small number of cores, as it reduces the effects of inter core communication, synchronization, and simplifies parallelization. In other words, it is more efficient to allocate a large number of small workers than a small number of bigger workers. Consequently, our experimentation uses the smallest CPU instances that can support sufficient RAM to hold the network. 

\begin{figure*}[h]
  \centering
  
  \includegraphics[width=340pt]{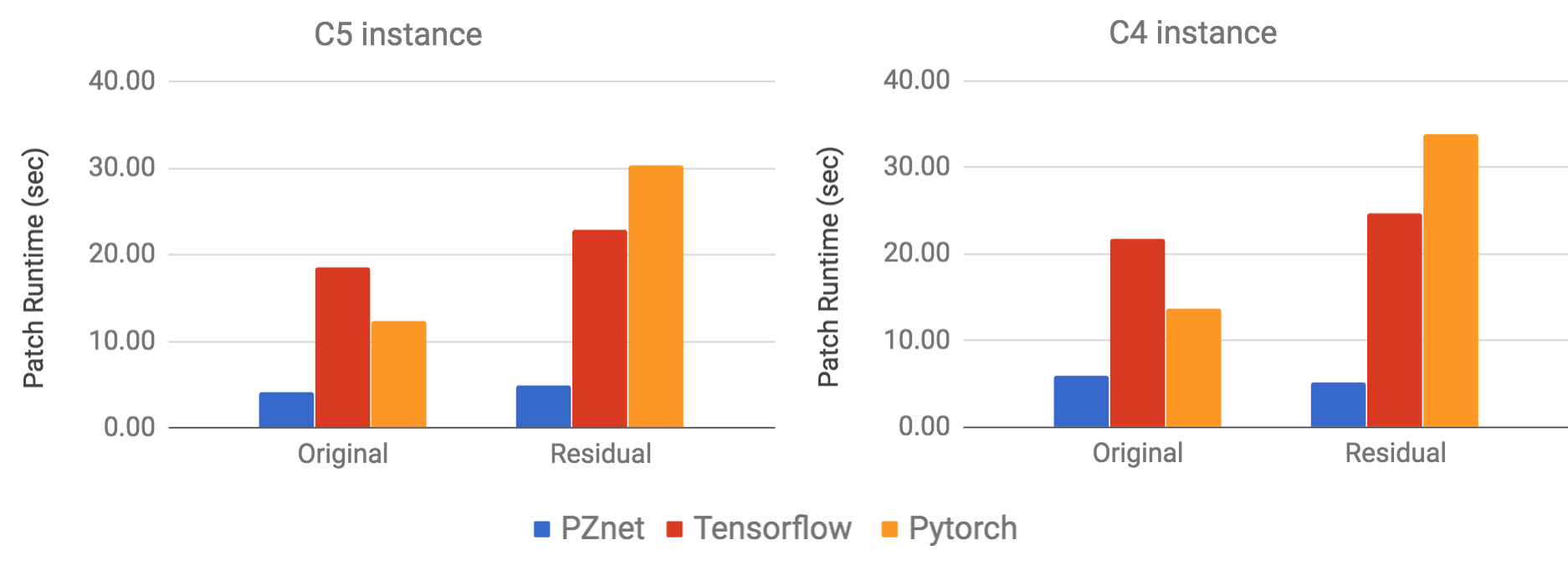}
  \caption{Comparison between CPU performance of PZnet, Tensorflow and PyTorch (lower is better).}
  \label{cpu_bar}
\end{figure*}

\subsection{Network Architectures}
The evaluation is done on 3 variants of 3D Unet, which is the state-of-the-art architecture for 3D image segmentation. The original 3D Unet uses convolutions with no padding and high numbers of feature maps. Several works propose using a modification of this architecture where padding was introduced in order to keep the input and output image size same \cite{lin2017feature}. Recent works \cite{drozdzal2016importance,quan2016fusionnet,lee2017superhuman} in 3D segmentation also include residual connections at each level of the UNet. For our evaluation, we use the original 3D unet as proposed by \cite{cciccek20163d}, a symmetric variation with padded convolutions, reduced number of features and addition instead of merging layers, and a residual variation used in \cite{lee2017superhuman}. These architectures will be referred to as Original, Symmetric, and Residual for the rest of this paper. As modern practices suggest, we include batch normalization and scaling layers after each convolution. We also use ELU as the activation function.

\begin{figure}[h]
  \centering
  \includegraphics[width=340pt]{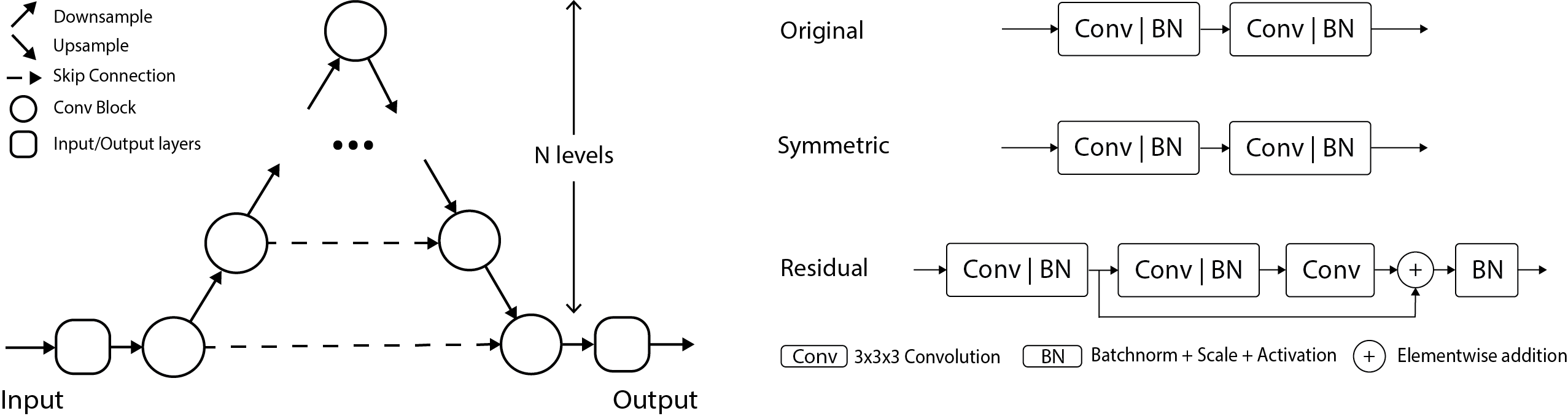}
  \caption{Left: generic Unet architecture with N-1 downsamples, upsamples and skip connections, Right: Convolutional block types for each model used in evaluation. BN corresponds to Batchnorm followed by Scale and Activation function }
  \label{unet}
\end{figure}

Fig.~\ref{unet} depicts the structure of Unet architecture families. Each network is composed of multiple levels of so-called convolutional blocks, connected to each other either through upsampling or downsampling layer (Fig.~\ref{unet} left). Convolutional blocks of Original and Symmetric variations are contain two consecutive convolutional layers, each followed by batch normalization and activation. Residual variation adds an additional residual connection and an additional convolutional layer to each block (Fig.~\ref{unet} right). 

The three architectures also differ in the number of featuremaps used for convolutions. The Original architecture uses 64 featurmaps for the convolutions on the first level, and doubles the number of convolutions on each consequent level. Symmetric architecture uses twice less featuremaps on each level. Reduced number of featuremaps is to compensate for the fact that Symmetric architecture uses convolution input paddding, and so the dimensions of each featuremap is higher. Residual architecture reduces the number of features even further, starting at 28 and going up in increments of 8 and 16 at each level, as described in \cite{lee2017superhuman}. 

For the timing experiments, weights were initialized with Xavier initialization \cite{glorot2010understanding} for the convolutional layers and as an identity for linear transformation layers. Each measurement was repeated for 60 iterations after a warm-up period of 10 iterations.

\subsection{CPU Performance}
In this experiment, we show that PZnet outperforms Tensorflow and Pytorch for 3D inference. Fig.~\ref{cpu_bar} compares CPU performance of PZnet, Tensorflow, and Pytorch. Symmetric architecture has the largest intermediate featuremaps, and the matrix multiplication 3D convolution implementation used in Tensorflow and Pytorch grows quadratically with featuremap sizes. This leads to RAM requirement that cannot be fulfilled with the low-core AWS instances used for experimentation. 

Tensorflow version used for evaluation was compiled with MKL, FMA and AVX2 support. Despite our best efforts, the master branch of Tensorflow could not be compiled with AVX512 support without producing segmentation faults during network inference. Tensorflow inference optimization scripts where used prior to timing. Pytorch implementation was also compiled with MKL and AVX2 support. The results show that PZnet outperforms Tensorflow by more than $3.4$x for all of the experiment settings, and outperforms Pytroch by more than $2.27$x. PZnet achieves maximum speedups over Tensorflow and Pytorch for Residual architecture. As will be shown later in this section, Residual architecture benefits most from the optimizations. 

\subsection{Cost Efficiency}

\subsubsection{Basic Blocks}
\begin{figure*}[h]
  \centering
  
  \includegraphics[width=340pt]{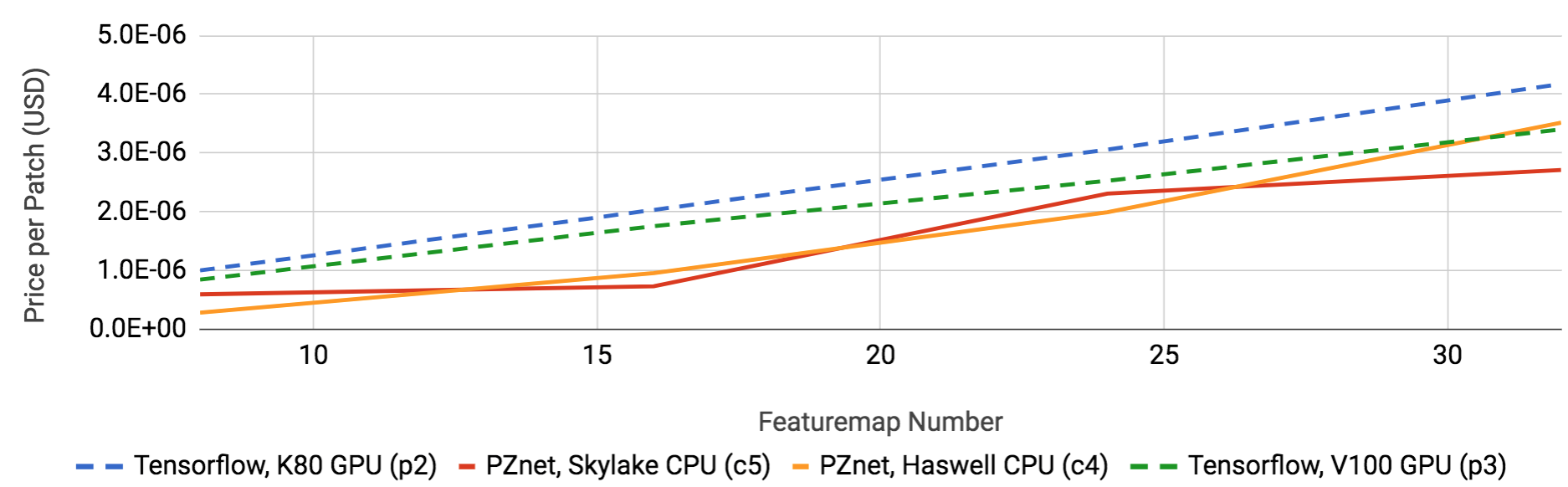}
  \caption{Cloud cost comparison for running 3D Unet basic block on various hardware  (lower is better). Dashed lines correspond to GPU inference with Tensorflow, solid lines correspond CPU inference with PZnet. Left: featuremap range 8 to 32. Right: featuremap range 8 to 64.}
  \label{3d_block}
\end{figure*}

In this experiment, we show that when convolutional feature number is small, CPU inference with PZnet can outperform GPU infererence in terms of cost efficiency. Inference of convolutional blocks with $8$ to $32$ featuremaps are timed on CPU and GPU instances. The results of the experiment are shown on Fig.~\ref{3d_block}. The $X$ axis corresponds to the number of featuremaps used in each of the convolutional layers. The $Y$ axis corresponds to the effective dollar price per patch. Price per patch is obtained by multiplying the execution time by the hourly price of the used AWS instance. The results show that when the featuremap number is small, CPUs can outperform GPUs in terms of price efficiency. In particular, when the featuremap number is 16, CPUs can be more than twice cheaper. GPUs outperformed CPUs for featuremap number greater than 32.

\subsubsection{Full Networks}
In this experiment, we study cost efficiency comparison between CPUs and GPUs on full network arcitectures. Three network architectures are tested overall: Original, Symmetric, and Residual variations of 3D Unet. The results of the experiment are presented in Table~\ref{table:cost}. Price per patch is obtained by multiplying the execution time by the hourly price of the used AWS instance. CPU inference is run with PZnet. CPU price per patch is taken to be the lowest among c4 and c5 instances for each data point. GPU price per patch are taken as the best of running Pytorch and Tensorflow accross p2 and p3 instances for each data point. 

For the networks with high number of featuremaps (Original, Symmetric), GPU cost per patch is lower than CPU cost per patch by factors of $2.76$x and $1.82$x. However, when the feature number becomes small (Residual) CPU cost efficiency is lower than GPU cost efficiency by a factor of $1.49$x. This confirms the hypothesis that CPU inference can be competitive with GPU inference in terms of cost efficiency when the featuremap number is low.

\begin{table}[ht]
\centering
\caption{Inference Cost Efficiency}
\label{table:cost}
\begin{tabular}{@{\extracolsep{4pt}}lccc}
\toprule   
{} & \multicolumn{3}{c}{Price Per Patch (USD)}\\
\cmidrule{1-4}
Platform & Original & Symmetric & Residual \\ 
CPU & 2.45E-05 & 4.47E-05 & 2.53E-05 \\ 
GPU & 8.86E-06 & 2.45E-05 & 3.77E-05 \\
\midrule
Best Platform & GPU   & GPU   & \textbf{CPU} \\
Margin        & 2.76x & 1.82x & \textbf{1.49x} \\
\bottomrule
\end{tabular}
\end{table}

\subsection{Optimizations}
The Table \ref{table:full_opt} presents the effects of the optimizations performed by PZnet. The original variation of 3D Unet benefits the least from the optimizations. The reason for that is two fold. First of all, original Unet uses the biggest numbers of featuremaps, which makes convolutions take bigger proportion of the total network runtime. Additional memory traversals and linear transformations take smaller proportion of the total inference time, and so their elimination is less critical for achieving high performance. Second, the Original variation of 3D does not use convolution input padding or element-wise addition. Thus, neither Output Padding Transformation nor Addition Fusion can be applied to the Original variation. Optimizations provide most benefits to the residual version of the UNet. This can be explained by the fact that residual UNet uses the smallest number of feature maps, which makes eliminated transforms and a large number of element-wise addition layers. 

\begin{table}[ht]
\centering
\caption{Full optimization speedup}
\label{table:full_opt}
\begin{tabular}{@{\extracolsep{4pt}}lccc}
\toprule   
{} & \multicolumn{3}{c}{ELU}\\
\cmidrule{2-4}
  
Instance   & Original & Symmetric & Residual \\ 
\midrule
c4          & 11.2\% & 14.9\% & 18.0\% \\ 
c5          & 9.4\%  & 17.2\%  & 20.0\%  \\
\bottomrule
\end{tabular}
\end{table}

\begin{figure}[h]
  \centering
  \includegraphics[width=200pt]{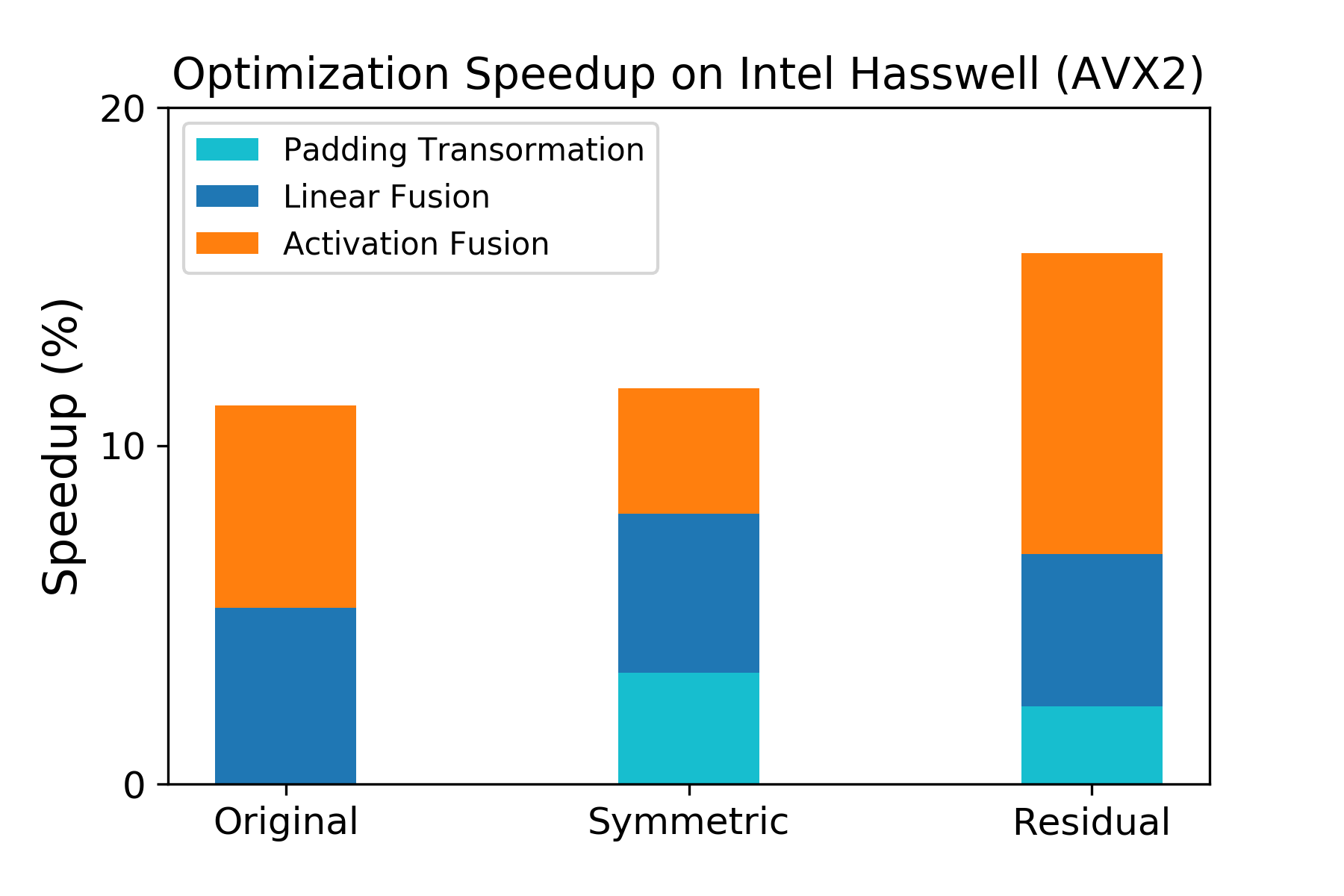}
  \caption{Effects of the optimization stages on the final performance. Measured on Intel Hasswell (AWS C4)}
  \label{bar_opts}
\end{figure}

Fig~\ref{bar_opts} breaks down the contribution of each optimization stage to the total speedup for each of the studied network architectures. Since Addition Fusion is used as an enabling optimization for Activation Fusion and Linear Fusion, its effects cannot be isolated. The results show that all three of the remaining optimizations, namely Liner Fusion, Activation Fusion and Padding Elimination contribute significantly to the total optimization speedup. The exception is the Original variation, which does not use convolution input padding, and so the Padding Elimination can have no effect. 

\section{Related Work}

Existing deep learning frameworks perform graph-level and memory allocation optimizations  \cite{sze2017efficient}. Theano \cite{bastien2012theano} and MXNet \cite{chen2015mxnet} implement arithmetic optimizations, such as folding of layers that multiply or add by scalar constants. Tensorflow \cite{abadi2016tensorflow} provides an option to freeze the network graph for inference. The optimizations performed by Tensorflow during freezing include removing train-only operations and remove nodes of the graph that are never reached given output node. 

Recently, Tensorflow introduced an optimization that fuses batch normalization weights into the convolution kernel, similar to PZnet.  NVIDIA Tesla RT\cite{tensorrt} provides fusions of convolution, bias and ReLU layers. However, the package does not provide 3D convolution support, which makes it inapplicable to 3D biomedical image processing. 

There has been several efforts to have CPU-efficient convolution implementations. ZNNi\cite{zlateski2016znni} combines CPU and GPU primitives for 3D ConvNet inference. Greater amount of RAM available on CPU workers allows CPU primitives to process bigger input volumes, which reduces the wasted computation at the volume border. Intel provides a Caffe branch\cite{intelcaffe} which utilizes MKL and MKLdnn libraries \cite{mkldnn}. Same as PZnet, Intel Caffe supports blocked data format. Unlike PZnet, Intel Caffe supports both training and inference. The crucial difference between PZnet and Caffe is that Intel Caffe and its underlying libraries do not support 3D convolutions.

\section{Conclusion}
We presented PZNet, an inference-only deep learning engine which is optimized for 3D inference on Intel Xeon CPUs. It utilizes ZNNPHi \cite{zlateski2017znnphi}, the state-of-the-art direct 3D convolution implementation. Additionally, we perform a series of inter-layer optimizations which reduce inference time by up to 20\% and evaluate on state-of-the-art dense prediction neural network architectures frequently used for biomedical image analysis.  

PZnet outperforms MKL-based \cite{intelcaffe} CPU implementations of PyTorch and Tesorflow by 3-8x. Moreover, we show PZNet CPU inference can be competitive with CUDnn based GPU inference in terms of cloud price efficiency. In the specific case of \cite{lee2017superhuman}, PZnet is able to outperform GPU inference in terms of cost efficiency by $50\%$.

\section{Acknoledgements}
This work has been supported by the Intelligence Advanced Research Projects Activity (IARPA) via Department of Interior/ Interior Business Center (DoI/IBC) contract number
D16PC0005. The U.S. Government is authorized to reproduce and distribute reprints
for Governmental purposes notwithstanding any copyright annotation thereon.
Disclaimer: The views and conclusions contained herein are those of the authors and
should not be interpreted as necessarily representing the official policies or
endorsements, either expressed or implied, of IARPA, DoI/IBC, or the U.S. Government. Additionally, this work was partially funded by TRI.
\bibliographystyle{abbrv}

\end{document}